\documentclass[conference]{IEEEtran}
\IEEEoverridecommandlockouts
\usepackage{cite}
\usepackage{amsmath,amssymb,amsfonts}
\usepackage{algorithmic}
\usepackage{graphicx}
\usepackage{textcomp}
\usepackage{xcolor}
\usepackage[raggedright]{sidecap}
\usepackage[singlelinecheck=false,justification=justified]{caption}
\usepackage{booktabs}
\usepackage{multirow}
\usepackage{enumitem}
\usepackage{balance}

\usepackage{caption}
\usepackage{subcaption}

\usepackage{sidecap}
\sidecaptionvpos{figure}{t}
\usepackage[colorlinks]{hyperref}
\usepackage{cite}

\def\BibTeX{{\rm B\kern-.05em{\sc i\kern-.025em b}\kern-.08em
    T\kern-.1667em\lower.7ex\hbox{E}\kern-.125emX}}
\begin{document}

\title{Face-voice Association in Multilingual Environments (FAME) Challenge 2024 Evaluation Plan}

\author{Muhammad Saad Saeed$^{1}$, Shah Nawaz$^{2}$, Muhammad Salman Tahir$^{1}$, Rohan Kumar Das$^{3}$, Muhammad Zaigham Zaheer$^{4}$,  \\Marta Moscati$^{2}$, Markus Schedl$^{2,6}$, Muhammad Haris Khan$^{4}$,  Karthik Nandakumar$^{4}$, Muhammad Haroon Yousaf$^{1}$  \\
$^{1}$Swarm Robotics Lab NCRA, University of Engineering and Technology Taxila, \\
$^{2}$Institute of Computational Perception, Johannes Kepler University Linz, Austria \\
$^{3}$Fortemedia Singapore, Singapore, \\
$^{4}$Mohamed bin Zayed University of Artificial Intelligence \\
$^{5}$Human-centered AI Group, AI Lab, Linz Institute of Technology, Austria \\
\tt \{mavceleb@gmail.com\}
}


\maketitle

\begin{abstract}
The advancements of technology have led to the use of multimodal systems in various real-world applications. Among them, the audio-visual systems are one of the widely used multimodal systems. In the recent years, associating  face and voice of a person has gained attention due to presence of unique correlation between them. The Face-voice Association in Multilingual Environments (FAME) Challenge 2024 focuses on exploring face-voice association under a unique condition of multilingual scenario. This condition is inspired from the fact that half of the world's population is bilingual and most often people communicate under multilingual scenario. The challenge uses a dataset namely, Multilingual Audio-Visual (MAV-Celeb) for exploring face-voice association in multilingual environments. This report provides the details of the challenge, dataset, baselines and task details for the FAME Challenge.

\end{abstract}

\section{Introduction}
\label{sec:intro}
\begin{figure}
    \centering
    \includegraphics[width=1\linewidth]{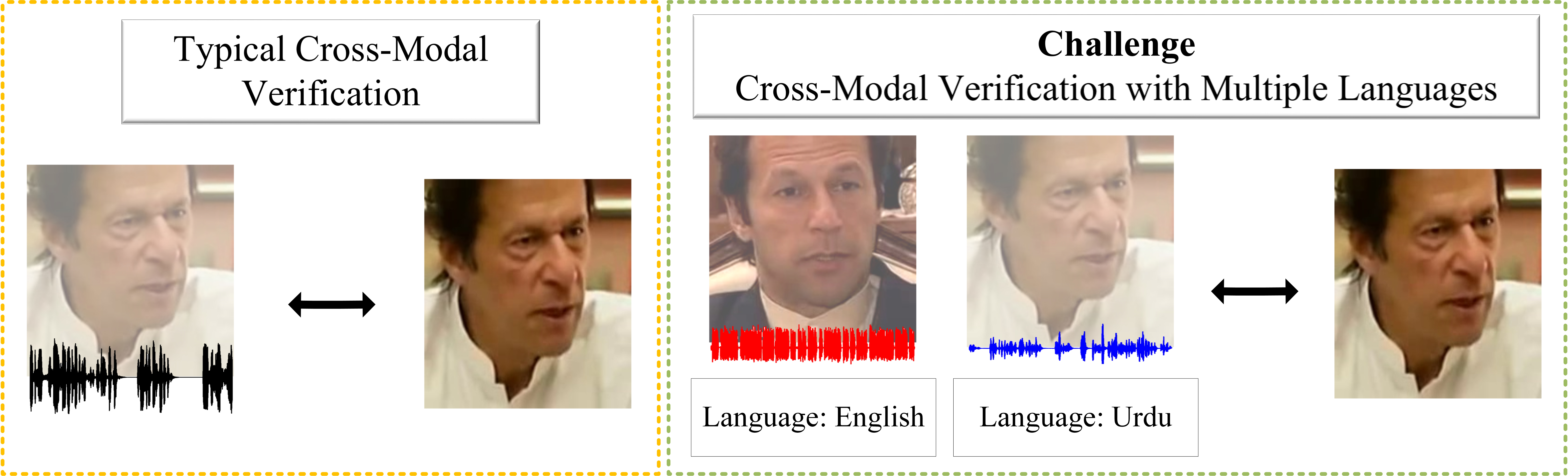}
    \caption{(Left) Standard Face-voice association is established with a cross-modal verification task. (Right) The FAME challenge 2024 extends the verification task to analyze the impact of multiple of languages.}
    \label{fig:verification+protocol}
\end{figure}

The face and voice of a person have unique characteristics and they are well used as biometric measures for person authentication either as a unimodal or multimodal~\cite{biometric_review,EUSIPCO2004,shah2023speaker}. A strong correlation has been found between face and voice of a person, which has attracted significant research interest~\cite{KAMACHI,nagrani2018seeing,FaceVoiceACM,tao20b_interspeech,saeed2022fusion,nawaz2019deep,saeed2023single,saeed2023single}. 
Though previous works have established association between faces and voices, none of these approaches investigated the effect of multiple languages on this task. As half of the population of world is bilingual and we are more often communicating in multilingual scenarios~\cite{bworld}, therefore, it is essential to investigate the effect of language for associating faces with the voices. 
Thus, the goal of the FAME challenge 2024 is to analyze the impact of multiple languages on face-voice association using cross-modal verification task. Fig.~\ref{fig:verification+protocol} provides an overview of the challenge.

In response, our prior research~\cite{nawaz2021cross} introduced a Multilingual Audio-Visual (MAV-Celeb) dataset to analyze the impact of language on face-voice association; it comprises of video and audio recordings of different celebrities speaking more than one language. For example, a celebrity named `Imran Khan' has audio information in Urdu and English languages.

\section{Challenge Objectives}

The FAME Challenge 2024 is planned with the primary objective to provide a common platform to academic and industrial researchers to develop and explore the impact of languages in face-voice association, which can be useful for various downstream tasks. The research conducted under this challenge is expected to spearhead the face-voice association task in one unique direction from the perspective of real-world scenarios. The challenge focuses on exploring the following, but not limited to:
\begin{itemize}
    \item To study the influence of language information in face-voice association 
    \item To explore language specific knowledge in face-voice association
    \item To explore language independent face-voice association models
    \item Any other related information with respect to language mismatch in face-voice association
\end{itemize}

\section{Dataset}
\label{sec:dataset}

\begin{figure}[t!]
    \centering
    \includegraphics[width=1\linewidth]{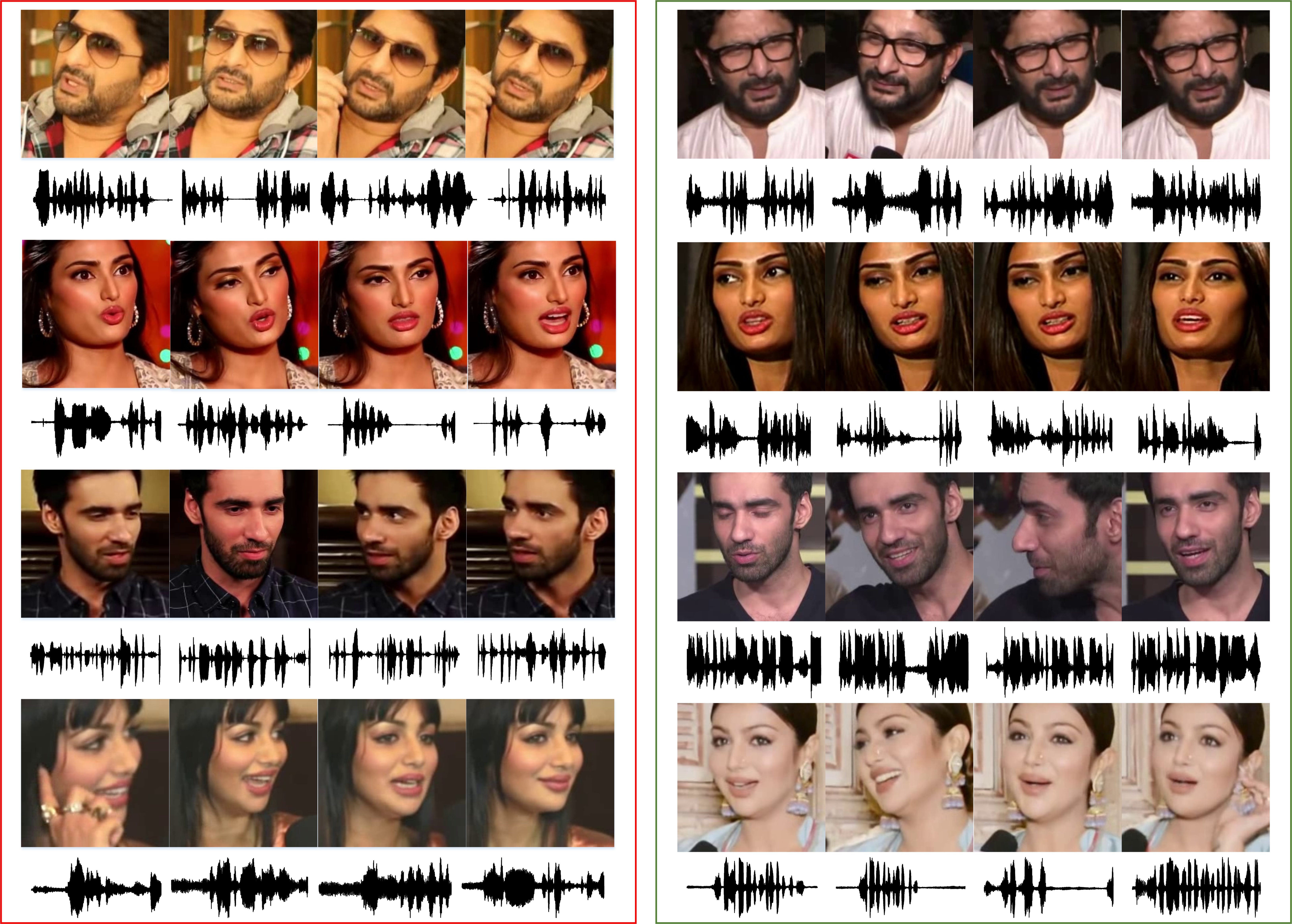}
    \caption{Audio-visual samples selected from MAV-Celeb dataset. The visual data contains various variations such as pose, lighting condition and motion. (Left) It contains information of celebrities speaking English and the (Right) block presents data of the same celebrity in Hindi.}
    \label{fig:mav-celeb}
\end{figure}
\begin{figure}
    \centering
    \includegraphics[width=0.5\linewidth]{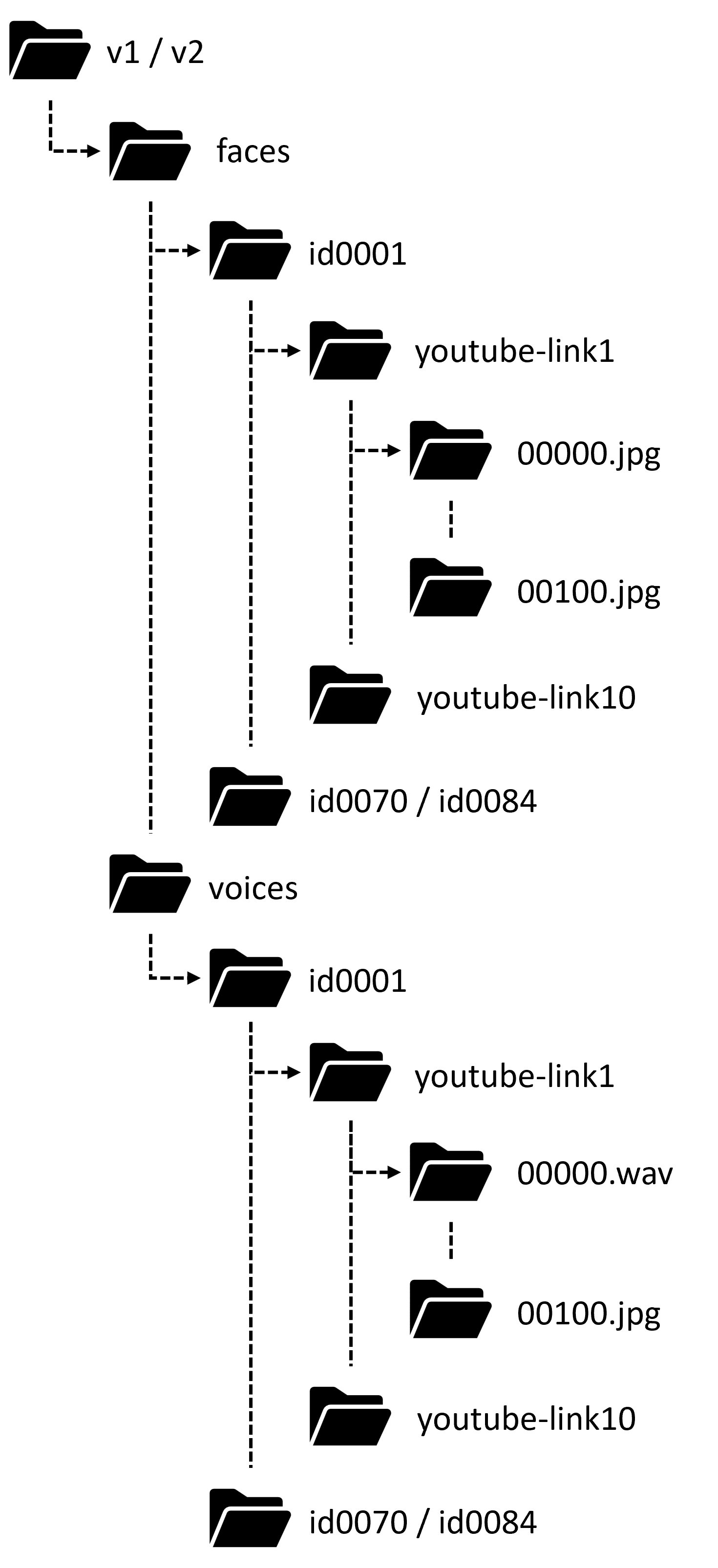}
    \caption{MAV-Celeb file structure.}
    \label{fig:file_structure}
\end{figure}

MAV-Celeb dataset provide data of $154$ celebrities in $3$ languages (English, Hindi, Urdu). 
These three languages have been selected because of several factors: i) They represent approximately 1.4 Billion bilingual/trilingual people; ii) The population is highly proficient in two or more languages; iii) There is a relevant corpus of different media that can be extracted from available online repositories (e.g. YouTube).
The collected videos cover a wide range of unconstrained, challenging multi-speaker environment including political debates, press conferences, outdoor interviews, quiet studio interviews, drama and movie clips.
Note that the visual data spans over a vast range of variations including poses, motion blur, background clutter, video quality, occlusions and lighting conditions. Moreover, videos are degraded with real-world noise like background chatter, music, overlapping speech and compression artifacts. 
Fig.~\ref{fig:mav-celeb} shows some audio-visual samples while Table~\ref{tab:data_stats} shows statistics of the dataset. 
The dataset contains $2$ splits English--Urdu (V$1$-EU) and English--Hindi (V$2$-EH) to analyze  performance measure across multiple languages. 
Fig.~\ref{fig:file_structure} shows the file structure of both splits. 
The pipeline followed in creating the dataset is available in our prior work~\cite{nawaz2021cross}.

\begin{table}[b]
\caption{Summary of MAV-Celeb dataset\label{tab:table1}}
\centering
\begin{tabular}{lccc}
\hline

{\bf Dataset} & {\bf E/U/V$1$-EU} & {\bf E/H/V$2$-EH}\\
\hline
\hline
\# of Celebrities & 70 & 84 \\
\hline
\# of male celebrities & 43 & 56 \\
\hline
\# of female celebrities & 27 & 28 \\
\hline
\# of videos & 402/555/957 & 646/484/1130 \\
\hline
\# of hours & 30/54/84 & 51/33/84 \\
\hline
\# of utterances & 6850/12706/19556 & 12579/8136/20715 \\
\hline
Avg \# of videos/celebrity & 6/8/14 & 8/6/14 \\
\hline
Avg \# of utterances/celebrity & 98/182/280 & 150/97/247 \\
\hline
Avg length of utterance & 15.8/15.3/15.6 & 14.6/14.6/14.6 \\
\hline
\end{tabular}
\label{tab:data_stats}
\end{table}

\section{Challenge Setup and Baseline} 
\label{sec:baseline}

\noindent \textbf{Challenge setup.} 
The MAV-Celeb dataset is divided into train and test splits consisting of disjoint identities from the same language typically known as unseen-unheard configuration~\cite{nagrani2018learnable}. Fig.~\ref{fig:verification+protocol} shows evaluation protocol at train and test time. At test time, the network is evaluated on a \textit{heard} and completely \textit{unheard} language on cross-modal verification task. The dataset splits V$1$-EU, V$2$-EH contains $64$–$6$, $78$–$6$ identities for train and test respectively. V$2$-EH split will be use in the progress phase. While, the final evaluation will be carried out in on test set of V$1$-EU.
Each line from test file in V$1$ and V$2$ has the following format. 

\begin{itemize}
    \item \texttt{ysuvkz41 voices/English/00000.wav faces/English/00000.jpg} 
    \item \texttt{tog3zj45 voices/English/00001.wav faces/English/00001.jpg}
    \item \texttt{ky5xfj1d voices/English/00002.wav faces/English/00002.jpg}
    \item \texttt{yx4nfa35 voices/English/01062.wav faces/English/01062.jpg}
    \item \texttt{bowsaf5e voices/English/01063.wav faces/English/01063.jpg}
\end{itemize}

\href{https://mavceleb.github.io/dataset}{MAV-Celeb} dataset is publicly available. We provided features from state-of-the-art methods representing faces and voices. Moreover, we created \href{https://mavceleb.github.io/dataset/competition.html}{FAME Challenge} website for additional information on the task.

\begin{figure*}
\centering
\includegraphics[scale=0.75]{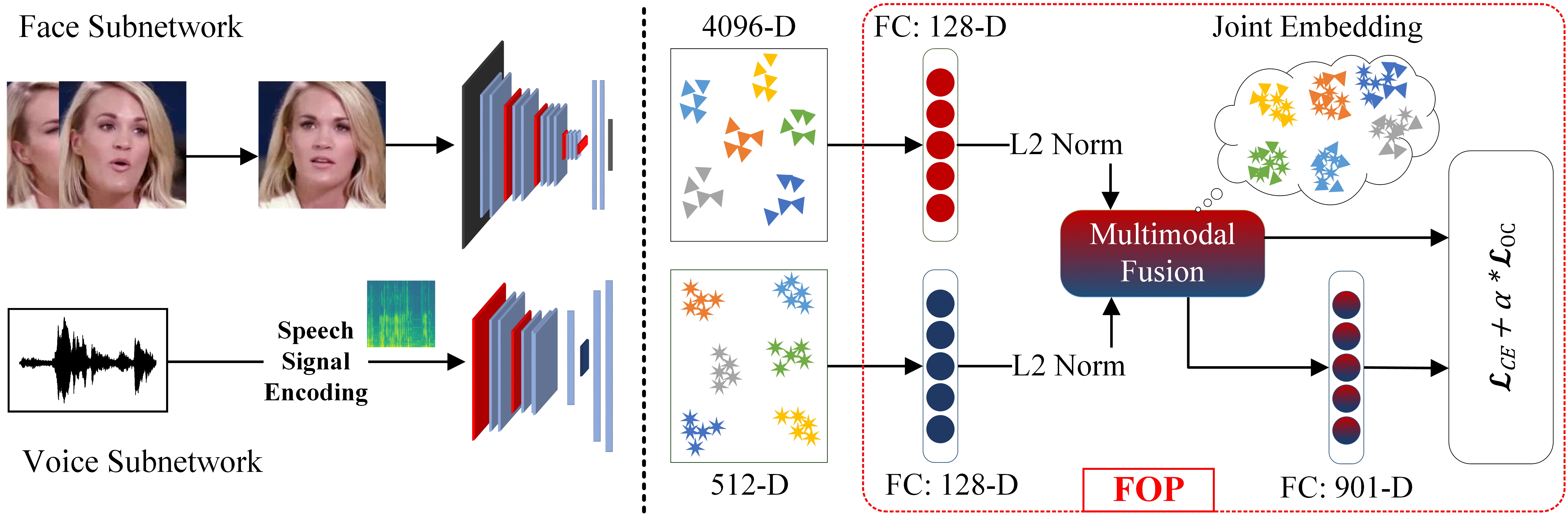}

   \caption{Overall architecture of our baseline method. Fundamentally, it is a two-stream pipeline which generates face and voice embeddings. We propose a light-weight, plug-and-play mechanism, dubbed as fusion and orthogonal projection (FOP) (shown in dotted red box). 
   } 
\label{fig:overall_framework_fusion}
\end{figure*}

\noindent \textbf{Evaluation Metric.} We are considering equal error rate (EER) as the metric for evaluating the challenge performance. We expect the challenge participants to submit a output score file for every test pairs to indicate how confident the system believes to have a match between the face and voice or in other words, the face and voice belongs to the same person. The higher the score is, the larger is the confidence of being the face and voice from the same person. In real-world applications, people may set a threshold to determine the if the pair belongs to same or different person as binary output. With the threshold higher, the false acceptance rate (FAR) will become lower, and the false rejection rate (FRR) will become higher. The EER is that optial point when both the errors FAR and FRR are equal. Therefore, EER becomes suitable to evaluate the performance of systems than the conventional accuracy since it independent of the threshold. Finally, the lower the EER it can characterize a better system.

\begin{table}[t]
\footnotesize
\caption{Cross-modal verification between faces and voices across multiple language on various test configurations of MAV-Celeb dataset. (EER: lower is better)}
\begin{center}
\begin{tabular}{llccc}
\hline
       &  & & \multicolumn{2}{c}{\textbf{V2-EH}}  \\
\hline\hline
Method & Configuration & Eng. test  & Hindi test & Overall score  \\
             & & (EER)                          & (EER)  & (EER) \\
\hline
\multirow{2}{*}{FOP~\cite{saeed2022fusion}}      & Eng. train     & 20.8   & 24.0 & \multirow{2}{*}{22.0} \\
                                                 & Hindi train     & 24.0           & 19.3 &  \\

\hline\hline
&  & & \multicolumn{2}{c}{\textbf{V1-EU}} \\
\hline
 &  & Eng. test  & Urdu test & Overall score   \\
             & & (EER)                          & (EER) &  (EER) \\
\hline
\multirow{2}{*}{FOP~\cite{saeed2022fusion}}      & Eng. train     & 29.3  & 37.9 & \multirow{2}{*}{33.4}  \\
                                                 & Urdu train     & 40.4  & 25.8 &  \\

\hline

\end{tabular}
\end{center}

\label{tab:lang}
\end{table}

\noindent \textbf{Baseline method.} The baseline method employ a two-stream pipeline to obtain the respective embeddings of both face and voice inputs. The first stream corresponds to a pre-trained convolutional neural network (CNN) on face modality~\cite{parkhi2015deep}. 
We take the penultimate layer's output of this CNN as the feature embeddings for an input face image. Likewise, the second stream is a pre-trained audio encoding network that outputs a feature embedding for an input audio signal (typically a short-term spectrogram)~\cite{xie2019utterance}. 
The baseline method exploits complementary cues from both modality embeddings to form enriched fused embeddings and imposes orthogonal constraints on them for learning discriminative joint face-voice embeddings, as shown in Fig~\ref{fig:overall_framework_fusion}. More information is available in our prior work~\cite{saeed2022fusion} and \href{https://github.com/mavceleb/mavceleb_baseline}{GitHub repository}.

\noindent \textbf{Baseline results.} Table~\ref{tab:lang} provides the baseline face-voice association results  with the impact of multiple languages on both splits of MAV-Celeb. The FAME challenge $2024$ encourages participants to explore novel ideas to improve performance on heard and unheard languages.


\section{Rules for System Development}

\begin{itemize}
    \item A pretrained model on heard language is allowed. 
    \item A pretrained model on unheard language is not allowed. The evaluation follows unheard-unseen and completely \textit{unheard} protocol. Each celebrity in the split has audio information in two languages; the model will be trained on one language (say Hindi) and then tested on heard language (Hindi) and completely \textit{unheard} language (English).
    \item The participants are required to submit a $2$ page system description in the ACM template to the challenge organizers. Teams without system description will be disqualified from the challenge. 
    \end{itemize}

\section{Registration Process}
The following Google Form to be used by participating teams for registration of their respective teams in the challenge.
\href{https://forms.gle/TfCfwgxkSL56iRmW9}{Registration form} 

\section{Submission of Results}
Within the directory containing the submission files, use zip archive.zip *.txt and do not zip the folder. Files should be named as:

\begin{itemize}
\item \texttt{sub\_score\_English\_heard.txt}
\item \texttt{sub\_score\_English\_unheard.txt}
\item \texttt{sub\_score\_Urdu\_heard.txt}
\item \texttt{sub\_score\_Urdu\_unheard.txt}
\end{itemize}

\begin{table*}[t]
\footnotesize
\caption{Configuration to analyze the impact of multiple languages on face-voice association with nomenclature.}
\begin{center}
\begin{tabular}{lcc}
\hline
          & \multicolumn{2}{c}{\textbf{V2-EH}}  \\
\hline\hline
 Configuration & Eng. test  & Hindi test\\
\hline
Eng. train     & \texttt{sub\_score\_English\_heard.txt}   & \texttt{sub\_score\_Hindi\_unheard.txt}  \\
Hindi train    & \texttt{sub\_score\_English\_unheard.txt} &  \texttt{sub\_score\_Hindi\_heard.txt}  \\

\hline\hline
  & \multicolumn{2}{c}{\textbf{V1-EU}} \\
\hline
& Eng. test  & Urdu test    \\
\hline
Eng. train     & \texttt{sub\_score\_English\_heard.txt} & \texttt{sub\_score\_Urdu\_unheard.txt}   \\
Urdu train     & \texttt{sub\_score\_English\_unheard.txt}  & \texttt{sub\_score\_Urdu\_heard.txt}  \\

\hline

\end{tabular}
\end{center}

\label{tab:lang_names}
\end{table*}

Moreover, Table~\ref{tab:lang_names} provides nomenclature of submission files.  For each file, we have kept the ground truth for fair evaluation during FAME challenge. Participants are expected to compute and submit text files including the \texttt{id} and \texttt{L2} scores in the following format:

\begin{itemize}
\item  \texttt{ysuvkz41 0.9988}
\item  \texttt{tog3zj45 0.1146}
\item  \texttt{ky5xfj1d 0.6514}
\item  \texttt{yx4nfa35 1.5321}
\item  \texttt{bowsaf5e 1.6578}
\end{itemize}

Each file is submitted through \href{https://codalab.lisn.upsaclay.fr/competitions/18534}{Codalab}. In the progress phase, each team will have $100$ submissions with maximum $10$ per day. While, we will overall $5$ submission in the evaluation phase. The overall score will be computed as:

\begin{equation}
\text { Overall Score }=(\text {Sum of all EERs}) / 4
\end{equation}

\section{Paper Submission}
The FAME 2024 challenge  is one of the Grand Challenges in ACM Multimedia 2024. The participants of the challenge are invited to grand challenge papers to ACM Multimedia following the official website.

\section{Important Dates}
Tentative timeline of the challenge following ACM Grand Challenge submission:

\begin{itemize}
\item Registration Period: 15 April-1 June 2024
\item Progress Phase: 15 April- 14 June 2024
\item Evaluation Phase: 15 June- 21 June 2024
\item Challenge Results: 27 June 2024
\item Submission of System Descriptions: 30 June 2024
\item ACM Grand Challenge Paper Submission: 24 July 2024
\end{itemize}


\balance
\bibliographystyle{IEEEbib}
\bibliography{IEEEbib}
\end{document}